\documentclass[runningheads]{llncs}
\usepackage{amsmath}
\usepackage{amssymb}
\usepackage{graphicx}
\usepackage{marvosym}
\usepackage{ulem}

\usepackage{color}
\usepackage{cite}
\usepackage{bm}
\usepackage{hyperref}
\usepackage{multirow}
\usepackage{tabularx}

\usepackage{mathtools}
\usepackage{subfigure}
\usepackage{url}
\usepackage[table]{xcolor}
\usepackage{lineno}
\usepackage{marginnote}
\usepackage{algorithm}
\usepackage{algpseudocode}
\usepackage{booktabs}
\usepackage[T1]{fontenc}
\makeatletter
\newcommand{\printfnsymbol}[1]{%
  \textsuperscript{\@fnsymbol{#1}}%
}

\begin{document}

\title{One-Shot Segmentation of Novel White Matter Tracts via Extensive Data Augmentation}
\titlerunning{One-Shot Novel WM Tract Segmentation via Data Augmentation}

\author{Wan Liu$^{\mathrm{1}}$, Qi Lu$^{\mathrm{1}}$, Zhizheng Zhuo$^{\mathrm{2}}$, Yaou Liu$^{\mathrm{2}}$, Chuyang Ye$^{\mathrm{1}}$\textsuperscript{(\Letter)}}
% index{Liu, Wan}
% index{Lu, Qi}
% index{Zhuo, Zhizheng}
% index{Liu, Yaou}
% index{Ye, Chuyang}
\authorrunning{Liu et al.}

\institute{$^{1}$School of Integrated Circuits and Electronics, Beijing Institute of Technology, Beijing, China\\
\email{chuyang.ye@bit.edu.cn}\\
$^{2}$Department of Radiology, Beijing Tiantan Hospital, Capital Medical University, Beijing, China\\}

\maketitle

\begin{abstract}
%The abstract should briefly summarize the contents of the paper in 15--250 words.
% \textit{White matter}~(WM) tract segmentation based on diffusion magnetic resonance imaging allows noninvasive identification of specific neural pathways. 
% Deep learning based methods have achieved state-of-the-art performance for automated WM tract segmentation, where segmentation models are trained with a large number of manually annotated scans accumulated throughout time.
Deep learning based methods have achieved state-of-the-art performance for automated \textit{white matter}~(WM) tract segmentation.
In these methods, the segmentation model needs to be trained with a large number of manually annotated scans, which can be accumulated throughout time.
When novel WM tracts---i.e., tracts not included in the existing annotated WM tracts---are to be segmented, additional annotations of these novel WM tracts need to be collected.
Since tract annotation is time-consuming and costly, it is desirable to make only a few annotations of novel WM tracts for training the segmentation model, and previous work has addressed this problem by transferring the knowledge learned for segmenting existing WM tracts to the segmentation of novel WM tracts. 
However, accurate segmentation of novel WM tracts can still be challenging in the one-shot setting, where only one scan is annotated for the novel WM tracts. 
In this work, we explore the problem of one-shot segmentation of novel WM tracts. 
% Not only do we exploit the knowledge transfer from the segmentation of existing WM tracts to the segmentation of novel WM tracts like existing methods, but we also transfer the knowledge from the generation of pseudo-labels of novel WM tracts.
% and to better address the problem of extremely scarce annotated data in the one-shot setting, we propose data augmentation and two-stage pretrain strategies.
Since in the one-shot setting the annotated training data is extremely scarce, based on the existing knowledge transfer framework, we propose to further perform extensive data augmentation for the single annotated scan, where synthetic annotated training data is produced. We have designed several different strategies that mask out regions in the single annotated scan for data augmentation.
To avoid learning from potentially conflicting information in the synthetic training data produced by different data augmentation strategies, we choose to perform each strategy separately for network training and obtain multiple segmentation models.
% Because there is not necessarily one strategy that is consistently better than the others, we choose to perform them separately for network training and obtain multiple segmentation models.
Then, the segmentation results given by these models are ensembled for the final segmentation of novel WM tracts.
% Then, the segmentation results obtained with these models are ensembled for the final segmentation of novel WM tracts.
Our method was evaluated on public and in-house datasets. The experimental results show that our method improves the accuracy of one-shot segmentation of novel WM tracts.

\end{abstract}

%
%Please note that the first paragraph of a section or subsection is not indented. The first paragraph that follows a table, figure,equation etc. does not need an indent, either.
\section{Introduction}
\label{sec:intro}

The segmentation of \textit{white matter}~(WM) tracts based on \textit{diffusion magnetic resonance imaging}~(dMRI) provides an important tool for the understanding of brain wiring~\cite{o2015does,yeatman2012tract}.
It allows identification of different WM pathways and has benefited various brain studies~\cite{stephens2020white,zarkali2020fiber}.
Since manual delineations of WM tracts can be time-consuming and subjective, automated approaches to WM tract segmentation are developed~\cite{wu2020tract,bazin2011direct,ye2015segmentation}, and methods based on \textit{convolutional neural networks}~(CNNs) have achieved state-of-the-art performance~\cite{zhang2020deep,wasserthal2018tractseg,liu2022volumetric}.
The success of CNN-based WM tract segmentation relies on a large number of annotated scans that are accumulated throughout time for network training.
However, in a new study, novel WM tracts that are not included in the existing annotated WM tracts may be of interest~\cite{toescu2021tractographic,macniven2020medial,banihashemi2021opposing} and need to be segmented. 
%However, there can also be novel WM tracts that are of interest in a new study, but they are not included in the existing annotated WM tracts~\cite{toescu2021tractographic,macniven2020medial,banihashemi2021opposing}. 
%However, there can also be novel WM tracts that are specific to a new study and not included in the existing annotated WM tracts~\cite{toescu2021tractographic,macniven2020medial,banihashemi2021opposing}. 
Repeating the annotation for the novel WM tracts on a large number of scans can be very laborious and prohibitive, and accurate segmentation of novel WM tracts becomes challenging when only a few annotations are made for them.
% Repeating the annotation for the novel WM tracts on a large number of scans can be very laborious and prohibitive, and accurate segmentation of novel WM tracts becomes challenging in the few-shot setting, where only a few annotations are made for them.

Previous work has addressed this few-shot segmentation problem with a transfer learning strategy, where the knowledge learned for segmenting existing WM tracts with abundant annotated data is transferred to the segmentation of novel WM tracts~\cite{lu2021knowledge}.
In~\cite{lu2021knowledge}, a CNN-based segmentation model pretrained for segmenting existing WM tracts is used to initialize the target network for segmenting novel WM tracts, so that even with only a few annotations of novel WM tracts the network can learn adequately for the segmentation during fine-tuning.
In addition, instead of using classic fine-tuning that discards the pretrained task-specific weights, an improved fine-tuning strategy is developed in~\cite{lu2021knowledge} for more effective knowledge transfer, where all weights in the pretrained model can be exploited for initializing the target segmentation model.
Despite the promising results achieved in~\cite{lu2021knowledge} for few-shot segmentation of novel WM tracts, when the number of annotated scans for novel WM tracts decreases to one, the segmentation is still challenging. Since fewer annotations are preferred to reduce the annotation time and cost, the development of accurate approaches to one-shot segmentation of novel WM tracts needs further investigation.

% {\color{red}Data augmentation has been used to increase the diversity of training data, which can be achieved with the basic image transformations, the generative models, or the image mixing and masking strategies~\cite{isensee2021nnu,ding2021modeling,yun2019cutmix,devries2017improved}. The basic transformations are always used online by CNN-based methods, such as spatial rotation and scaling~\cite{isensee2021nnu}. The training of generative models usually requires a large amount of annotated data, and unannotated data is needed for few-shot or one-shot settings~\cite{ding2021modeling}. Image mixing requires at least two annotated images~\cite{yun2019cutmix}. Therefore, image masking is most suitable for our one-shot problem~\cite{devries2017improved}.}

In this work, we seek to improve one-shot segmentation of novel WM tracts. 
We focus on volumetric WM tract segmentation~\cite{wasserthal2018tractseg,liu2022volumetric}, where voxels are directly labeled without necessarily performing tractography~\cite{basser2000vivo}.
Since in the one-shot setting annotated training data is extremely scarce, based on the pretraining and fine-tuning framework developed in~\cite{lu2021knowledge}, we propose to address the one-shot segmentation problem with extensive data augmentation. 
Existing data augmentation strategies can be categorized into those based on basic image transformation~\cite{isensee2021nnu,wasserthal2018tractseg}, generative models\cite{ding2021modeling}, image mixing~\cite{yun2019cutmix,zhang2021carvemix}, and image masking~\cite{devries2017improved}. 
Basic image transformation is already applied by default in CNN-based WM tract segmentation~\cite{wasserthal2018tractseg,liu2022volumetric}, yet it is insufficient for the one-shot segmentation due to the limited diversity of the augmented data. 
The training of generative models usually requires a large amount of annotated data, or at least a large amount of unannotated data~\cite{ding2021modeling}, which is not guaranteed in the one-shot setting. 
Image mixing requires at least two annotated images~\cite{zhang2021carvemix}, which is also infeasible in the one-shot setting.
Therefore, we develop several strategies based on image masking for data augmentation, where the single annotated image is manipulated by masking out regions in different ways to synthetize additional training data.

The masking is performed randomly either with uniform distributions or according to the spatial location of novel WM tracts, and the annotation of the synthetic image can also be determined in different ways.
The augmented data is used to fine-tune the model for segmenting novel WM tracts.
To avoid learning from potentially conflicting information in the synthetic data produced by different strategies, we choose to perform each data augmentation strategy separately to train multiple segmentation models, and the outputs of these models are ensembled for the final segmentation.
We evaluated the proposed method on two brain dMRI datasets. The results show that our method improves the accuracy of one-shot segmentation of novel WM tracts. The code of our method is available at \url{https://github.com/liuwan0208/One-Shot-Extensive-Data-Augmentation}.

\section{Methods}
\label{sec:method}

\subsection{Background: Knowledge Transfer for Segmenting Novel WM Tracts}
\label{sec:tl}

% Suppose we are given a CNN-based segmentation model that segments existing WM tracts, for which a large number of annotations have been accumulated for network training.
% We are interested in segmenting novel WM tracts for which only one scan is annotated due to the annotation cost.
Suppose we are given a CNN-based model $\mathcal{M}_{\rm{e}}$ that segments $M$ existing WM tracts, for which a large number of annotations have been accumulated for training.
We are interested in training a CNN-based model $\mathcal{M}_{\rm{n}}$ for segmenting $N$ novel WM tracts, for which only one scan is annotated due to annotation cost.

Existing work~\cite{lu2021knowledge} has attempted to address this problem with a transfer learning strategy based on the pretraining and fine-tuning framework, where $\mathcal{M}_{\rm{e}}$ and $\mathcal{M}_{\rm{n}}$ share the same network structure for feature extraction, and their last task-specific layers are different.
In classic fine-tuning, the network weights of the learned feature extraction layers of $\mathcal{M}_{\rm{e}}$ are used to initialize the feature extraction layers of $\mathcal{M}_{\rm{n}}$, and the task-specific layer of $\mathcal{M}_{\rm{n}}$ is randomly initialized.
Then, all weights of $\mathcal{M}_{\rm{n}}$ are fine-tuned with the single scan annotated for novel WM tracts.
However, the classic fine-tuning strategy discards the information in the task-specific layer of $\mathcal{M}_{\rm{e}}$. As different WM tracts cross or overlap, existing and novel WM tracts can be correlated, and the task-specific layer of $\mathcal{M}_{\rm{e}}$ for segmenting existing WM tracts may also bear relevant information for segmenting novel WM tracts.
Therefore, to exploit all the knowledge learned in $\mathcal{M}_{\rm{e}}$, in~\cite{lu2021knowledge} an improved fine-tuning strategy is developed, which, after derivation, can be conveniently achieved with a warmup stage. Specifically, the feature extraction layers of $\mathcal{M}_{\rm{n}}$ are first initialized with those of $\mathcal{M}_{\rm{e}}$.
Then, in the warmup stage, the feature extraction layers of $\mathcal{M}_{\rm{n}}$ are fixed and only the last task-specific layer of $\mathcal{M}_{\rm{n}}$ (randomly initialized) is learned with the single annotated image. Finally, all weights of $\mathcal{M}_{\rm{n}}$ are jointly fine-tuned with the single annotated image.

\subsection{Extensive Data Augmentation for One-Shot Segmentation of Novel WM Tracts}
\label{sec:daug}

Although the transfer learning approach in~\cite{lu2021knowledge} has improved the few-shot segmentation of novel WM tracts, when the training data for novel WM tracts is extremely scarce with only one annotated image, the segmentation is still challenging. 
% Although the transfer learning approach in~\cite{lu2021knowledge} has improved the segmentation of novel WM tracts in the few-shot setting, when the training data for novel WM tracts is extremely scarce with only one annotated image, the segmentation is still challenging. 
Therefore, we continue to explore the problem of one-shot segmentation of novel WM tracts.
Based on the pretraining and fine-tuning framework developed in~\cite{lu2021knowledge}, we propose to more effectively exploit the information in the single annotated image with extensive data augmentation for network training.
% for network fine-tuning.
% Mathematically, suppose the annotated image is $\mathbf{X}$ and its annotation is $\mathbf{Y}$ (0 for background and 1 for foreground); then we seek to obtain a set of synthetic annotated training images $\widetilde{\mathbf{X}}$ and the synthetic annotations $\widetilde{\mathbf{Y}}$ by transforming $\mathbf{X}$ and $\mathbf{Y}$. 
Suppose the annotated image is $\mathbf{X}$ and its annotation is $\mathbf{Y}$ (0 for background and 1 for foreground); then we obtain a set of synthetic annotated training images $\widetilde{\mathbf{X}}$ and the synthetic annotations $\widetilde{\mathbf{Y}}$ by transforming $\mathbf{X}$ and $\mathbf{Y}$. 
We develop several data augmentation strategies for the purpose, which are described below.

\subsubsection{\textbf{Random Cutout}}
First, motivated by the Cutout data augmentation method~\cite{devries2017improved} that has been successfully applied to image classification problems, we propose to obtain the synthetic image $\widetilde{\mathbf{X}}$ by transforming $\mathbf{X}$ with region masking:
\begin{eqnarray}
\widetilde{\mathbf{X}} &=& \mathbf{X}\odot (1-\mathbf{M}),
\label{eq:x}
\end{eqnarray}
where $\odot$ represents voxelwise multiplication and $\mathbf{M}$ is a binary mask representing the region that is masked out.
%where $\odot$ represents voxelwise multiplication and $\mathbf{M}$ is a binary mask that represents the region that is masked out.
$\mathbf{M}$ is designed as a 3D box randomly selected with uniform distributions. Mathematically, suppose the ranges of $\mathbf{M}$ in the $x$-, $y$-, and $z$-direction are $(r_x, r_x + w_x)$, $(r_y, r_y + w_y)$, and $(r_z, r_z + w_z)$, respectively; then we follow \cite{yun2019cutmix} and select the box as
%$\mathbf{M}$ is designed as a 3D box randomly selected with uniform distributions. Mathematically, suppose the ranges of $\mathbf{M}$ in the $x$-, $y$-, and $z$-direction are $(r_x, r_x + r_w)$, $(r_y, r_y + r_h)$, and $(r_z, r_z + r_d)$, respectively; then we follow \cite{yun2019cutmix} and select the box as
\begin{eqnarray}
\label{eq:rc1}
&&r_x\sim{U(0,R_x)},\, r_y\sim{U(0,R_y)}, \,r_z\sim{U(0,R_z)},\\
&&w_x=R_x\sqrt{1-\lambda}, \, w_y=R_y\sqrt{1-\lambda}, \, w_z=R_z\sqrt{1-\lambda}, \,\lambda \sim \rm{Beta}(1, 1),
\label{eq:rc2}
\end{eqnarray}
where $U(\cdot, \cdot)$ represents the uniform distribution, $R_x$, $R_y$, and $R_z$ are the image dimensions in the $x$-, $y$-, and $z$-direction, respectively, and $\lambda$ is sampled from the beta distribution $\rm{Beta}(1, 1)$ to control the size of the masked region.

The voxelwise annotation $\widetilde{\mathbf{Y}}$ for $\widetilde{\mathbf{X}}$ also needs to be determined. Intuitively, we can obtain $\widetilde{\mathbf{Y}}$ with the same masking operation for $\widetilde{\mathbf{X}}$:
\begin{eqnarray}
\widetilde{\mathbf{Y}} &=& \mathbf{Y}\odot (1-\mathbf{M}). 
\label{eq:y1}
\end{eqnarray}
The strategy that obtains synthetic training data using Eqs. (\ref{eq:x}) and (\ref{eq:y1}) with the sampling in Eqs.~(\ref{eq:rc1}) and (\ref{eq:rc2}) is referred to as \textit{Random Cutout One}~(RC1).
Besides RC1, it is also possible to keep the original annotation $\mathbf{Y}$ for the masked image $\widetilde{\mathbf{X}}$, so that the network learns to restore the segmentation result in the masked region. In this case, the synthetic annotation $\widetilde{\mathbf{Y}}$ is simply determined as
\begin{eqnarray}
\widetilde{\mathbf{Y}} &=& \mathbf{Y}.
\label{eq:y2}
% \mathbf{X}_i = \mathbf{X}\odot (1-\mathbf{M}), \mathbf{Y}_i=\mathbf{Y}\odot (1-\mathbf{M}),
\end{eqnarray}
The use of Eqs. (\ref{eq:x}) and (\ref{eq:y2}) for obtaining synthetic training data with the sampling in Eqs. (\ref{eq:rc1}) and (\ref{eq:rc2}) is referred to as \textit{Random Cutout Two}~(RC2).

\subsubsection{\textbf{Tract Cutout}} Since we perform data augmentation for segmenting novel WM tracts, in addition to RC1 and RC2, it is possible to obtain $\mathbf{M}$ with a focus on the novel WM tracts.
To this end, we design the computation of $\mathbf{M}$ as 
\begin{eqnarray}
\mathbf{M}=\left \lceil \frac{1}{N}  \sum_{j=1}^{N}{a^j\mathbf{Y}^j}
\right \rceil,
\label{eq:tc}
\end{eqnarray}
where $\mathbf{Y}^j$ denotes the annotation of the $j$-th novel WM tract in $\mathbf{Y}$,  $a^j$ is sampled from the Bernoulli distribution $\mathrm{Bernoulli}(0.5)$ to determine whether $\mathbf{Y}^j$ contributes to the computation of $\mathbf{M}$, and $\lceil\cdot\rceil$ represents the ceiling operation. 
% where $\mathbf{Y}^j$ denotes the annotation mask of $j$-th novel WM tract in $\mathbf{Y}$,  $a^j$ is sampled from the Bernoulli distribution $\mathrm{Bernoulli}(0.5)$ and determines whether the $j$-th novel WM tract contributes to the computation of $\mathbf{M}$, and $\lceil\cdot\rceil$ represents the ceiling operation. 
In this way, $\mathbf{M}$ is the union of the regions of a randomly selected subset of the novel WM tracts, and thus the masked region depends on the novel WM tracts.

With the masking strategy in Eq.~(\ref{eq:tc}),  we can still use Eq.~(\ref{eq:y1}) or (\ref{eq:y2}) to determine the synthetic annotation.
When Eq.~(\ref{eq:y1}) or (\ref{eq:y2}) is used, the data augmentation strategy is named \textit{Tract Cutout One}~(TC1) or \textit{Tract Cutout Two}~(TC2), respectively.
No duplicate synthetic images are allowed in TC1 or TC2.
% For convenience, we refer to the combination of Eqs.~(\ref{eq:x}) and (\ref{eq:y1}) with the sampling in Eq.~(\ref{eq:tc}) as \textit{Tract Cutout One}~(TC1), and we refer to the use of Eqs.~(\ref{eq:x}) and (\ref{eq:y1}) with the sampling in Eq.~(\ref{eq:tc}) as \textit{Tract Cutout Two}~(TC2).

% \subsection{Network Training with Augmented Data}

\subsubsection{Network Training with Augmented Data}

% The proposed data augmentation strategies are integrated with the improved fine-tuning framework developed in~\cite{lu2021knowledge}.
% First, the feature extraction layers---i.e., all layers before the last task-specific layer---of $\mathcal{M}_{\rm{n}}$ are initialized with those in $\mathcal{M}_{\rm{e}}$ that has been well trained with abundant annotations of existing WM tracts.
By repeating the region masking in each data augmentation strategy, a set of synthetic annotated images can be produced.
% By repeating the region masking in each data augmentation strategy, a set of synthetic annotated images can be produced for the fine-tuning of $\mathcal{M}_{\rm{n}}$.
% The synthetic data is used for the fine-tuning of $\mathcal{M}_{\rm{n}}$.
Since the synthetic images can appear unrealistic, they are used only in the warmup stage of the improved fine-tuning framework in~\cite{lu2021knowledge}, where the last layer of $\mathcal{M}_{\rm{n}}$ is learned.
% Since the appearance of the synthetic data can be unrealistic, the synthetic training data is used together with the real annotated training scan only in the warmup stage of the improved fine-tuning framework in~\cite{lu2021knowledge}, where the last layer of $\mathcal{M}_{\rm{n}}$ is learned.
In the final fine-tuning step that updates all network weights in $\mathcal{M}_{\rm{n}}$, only the real annotated training image is used.
In addition, to avoid that the network learns from potentially conflicting information in the synthetic data produced by different strategies, we choose to perform each data augmentation strategy separately and obtain four different networks for segmenting novel WM tracts.
% In addition, as there is not necessarily a data augmentation strategy that is better than the other three, we choose to perform each of them separately for network fine-tuning and obtain four different networks for segmenting novel WM tracts.
At test time, the predictions of the four networks are ensembled with majority voting\footnote{The tract label is set to one when the votes are tied.} to obtain the final segmentation.

\subsection{Implementation Details}
\label{sec:detail}

We use the state-of-the-art TractSeg architecture~\cite{wasserthal2018tractseg} for volumetric WM tract segmentation as our backbone segmentation network, which takes fiber orientation maps as input.  
Like~\cite{wasserthal2018tractseg}, the fiber orientation maps are computed with \textit{constrained spherical deconvolution}~(CSD)~\cite{tournier2007robust} or \textit{multi-shell multi-tissue CSD}~(MSMT-CSD)~\cite{jeurissen2014multi} for single-shell or multi-shell dMRI data, respectively, and three fiber orientations are allowed in the network input~\cite{wasserthal2018tractseg}.
% Like~\cite{wasserthal2018tractseg}, the fiber orientation maps are computed from dMRI with \textit{constrained spherical deconvolution}~(CSD)~\cite{tournier2007robust} or \textit{multi-shell multi-tissue CSD}~(MSMT-CSD)~\cite{jeurissen2014multi} implemented in MRtrix3~\cite{tournier2019mrtrix3} for single-shell or multi-shell dMRI data, respectively. 
% Also, to improve the generalization to dMRI data acquired with different protocols, we follow ~\cite{wasserthal2018tractseg} and generate three different types of fiber orientation maps are computed from pretraining dMRI scan with MSMT-CSD as in~\cite{wasserthal2018tractseg}, which are randomly sampled during network training. 
% Three fiber orientations are allowed in the network input and thus the number of input channels is nine~\cite{wasserthal2018tractseg}.
We follow~\cite{wasserthal2018tractseg} and perform 2D WM tract segmentation for each image view separately, and the results are fused to obtain the final 3D WM tract segmentation.

The proposed data augmentation is performed offline.
Since given $N$ novel WM tracts TC1 or TC2 can produce at most $2^N-1$ different images, we set the number of synthetic scans produced by each data augmentation strategy to $\min(2^N-1,100)$. 
Note that traditional data augmentation, such as elastic deformation, scaling, intensity perturbation, etc., is applied online in TractSeg to training images. Thus, these operations are also applied to the synthetic training data online. 
The training configurations are set according to TractSeg, where Adamax~\cite{kingma2014adam} is used to minimize the binary cross entropy loss with a batch size of 56, an initial learning rate of 0.001, and 300 epochs~\cite{wasserthal2018tractseg}. 
The model corresponding to the epoch with the best segmentation accuracy on a validation set is selected.
% Model selection is performed based on the segmentation accuracy on a validation set.
% The best model is selected according to the segmentation accuracy on a validation set.

\section{Experiments}
\subsection{Data Description and Experimental Settings}

For evaluation, experiments were performed on the publicly available \textit{Human Connectome Project}~(HCP) dataset~\cite{van2013wu} and an in-house dataset. 
The dMRI scans in the HCP dataset were acquired with 270 diffusion gradients (three $b$-values) and a voxel size of 1.25 mm isotropic. 
In~\cite{wasserthal2018tractseg} 72 major WM tracts were annotated for the HCP dataset, and the annotations are also publicly available. 
% In~\cite{wasserthal2018tractseg} 72 major WM tracts were annotated for the HCP dataset, and the annotations are available at \url{https://doi.org/10.5281/zenodo.1088277}. 
For the list of the 72 WM tracts, we refer readers to~\cite{wasserthal2018tractseg}.
The dMRI scans in the in-house dataset were acquired with 270 diffusion gradients (three $b$-values) and a voxel size of 1.7 mm isotropic.
In this dataset, only ten of the 72 major WM tracts were annotated due to the annotation cost. 
% In this dataset, only ten of the 72 major WM tracts were annotated due to the annotation cost. 

Following~\cite{lu2021knowledge}, we selected the same 60 WM tracts as existing WM tracts, and a segmentation model was pretrained for these tracts with the HCP dMRI scans, where 48 and 15 annotated scans were used as the training set and validation set, respectively. 
To evaluate the performance of one-shot segmentation of novel WM tracts, we considered a more realistic and challenging scenario where novel WM tracts are to be segmented on dMRI scans that are acquired differently from the dMRI scans annotated for existing WM tracts.
Specifically, instead of using the original HCP dMRI scans for segmenting novel WM tracts, we generated clinical quality scans from them. The clinical quality scans were generated by selecting only 34 diffusion gradients at $b$ = 1000 s/mm$^{2}$ and downsampling the selected diffusion weighted images in the spatial domain by a factor of two to the voxel size of 2.5~mm isotropic.
% to a resolution of 2.5~mm isotropic
%and the tract annotations were also downsampled accordingly.
The tract annotations were also downsampled accordingly.
Since the dMRI scans in the in-house dataset were acquired differently from the original HCP dMRI scans, they were directly used for evaluation together with their original annotations.
\begin{table}[!t]
\caption{A summary of the cases (CQ1, CQ2, CQ3, IH1, IH2, and IH3) of novel WM tracts considered in the experiments and the tract abbreviations (abbr.). The checkmark (\checkmark)~indicates that the tract was included in the case.}
\centering
\resizebox{0.95\columnwidth}{!}{
% \begin{tabular}{l l  c c c c c c}
\begin{tabular}{>{\arraybackslash}p{5.5cm} l >{\centering\arraybackslash}p{1.0cm} >{\centering\arraybackslash}p{0.8cm} >{\centering\arraybackslash}p{0.8cm} >{\centering\arraybackslash}p{0.8cm}
%\begin{tabular}{l l >{\centering\arraybackslash}p{1.0cm} >{\centering\arraybackslash}p{0.8cm} >{\centering\arraybackslash}p{0.8cm} >{\centering\arraybackslash}p{0.8cm}
>{\centering\arraybackslash}p{0.8cm}
>{\centering\arraybackslash}p{0.8cm}}
\hline
\hline
WM Tract Name&Abbr. & CQ1 & CQ2 &CQ3& IH1 & IH2 & IH3\\
\hline
Corticospinal tract left & CST\_left& \checkmark &\checkmark  & \checkmark & \checkmark &\checkmark  & \checkmark \\ 
Corticospinal tract right &CST\_right& \checkmark &\checkmark  & \checkmark & \checkmark &\checkmark  & \checkmark  \\
Fronto-pontine tract left & FPT\_left&  &  & \checkmark & & & \checkmark  \\ 
Fronto-pontine tract right & FPT\_right&  &  & \checkmark & & & \checkmark  \\ 
Inferior longitudinal fascicle left & ILF\_left& & & \checkmark& & & \\
Inferior longitudinal fascicle right & ILF\_right& & & \checkmark& & & \\
Optic radiation left & OR\_left & \checkmark& \checkmark&\checkmark & \checkmark& \checkmark & \checkmark \\
Optic radiation right & OR\_right&\checkmark &\checkmark &\checkmark & \checkmark& \checkmark& \checkmark\\
Parieto-occipital pontine tract left & POPT\_left& & \checkmark&\checkmark & & \checkmark&\checkmark  \\ 
Parieto-occipital pontine tract right &POPT\_right & &\checkmark & \checkmark& & \checkmark&\checkmark\\ 
Uncinate fascicle left & UF\_left& & &\checkmark & & &\checkmark\\
Uncinate fascicle right & UF\_right& & &\checkmark & & &\checkmark \\
\hline
\hline
\end{tabular}}
\label{tab:tract_list}
\end{table}

% CQ and IH were used to evaluate the accuracy of one-shot segmentation of novel WM tracts separately based on the model pretrained on the original HCP dataset for segmenting existing WM tracts. 
The two datasets were used to evaluate the accuracy of one-shot segmentation of novel WM tracts separately based on the model pretrained on the original HCP dataset for segmenting existing WM tracts. 
We considered three cases of novel WM tracts for each dataset. For the clinical quality scans, the three cases are referred to as CQ1, CQ2, and CQ3, and for the in-house dataset, the three cases are referred to as IH1, IH2, and IH3. The details about these cases are summarized in Table~\ref{tab:tract_list}.
For each dataset, only one scan was selected from it for network training, together with the corresponding annotation of novel WM tracts.\footnote{The single annotated scan was also used as the validation set for model selection.}
For the clinical quality dataset and the in-house dataset, 30 and 16 scans were used for testing, respectively, where the annotations of novel WM tracts were available and only used to measure the segmentation accuracy.

% For network finetuning , the other 12 tracts of HCP dataset are used  as novel WM tracts as in ~\cite{lu2021knowledge}, which are listed in Table~\ref{tab:tract_list}. To show the performance of our method in clinical quality dMRI data, like~\cite{lu2021volumetric, liu2022volumetric}, we generate low-quality scans from the original scans in HCP dataset by selecting 34 gradient directions at  $b$ = 1000 s/mm$^{2}$ and downsampling (a factor of two)  these slices to 2.5mm isotropic resolution in spatial domain, and the annotations are also downloaded in spatial domain. This dataset is referred to as HCP\_clinical. One generated scan is used as the single annotated scans for novel WM tract, and 30 scans are used for test.  

% In addition, like~\cite{liu2022volumetric},  we utilize dMRI scans in the private dataset acquired at Beijing Tiantan Hospital for evaluation. The dMRI scans were acquired with 270 diffusion gradient. The resolution is 1.7 mm isotropic, and the image dimension is $128\times128\times80$. Considering the cost of annotation, ten tracts listed in Table~\ref{tab:tract_list} in bold are manually delineated~\cite{liu2022volumetric}, which are used as the novel WM tracts for the private scans. To decrease the data quality, we selecting 36 gradient directions at 18 $b$ = 1000 s/mm$^{2}$ and 18 $b$ = 1000 s/mm$^{2}$. This dataset is referred to as BT\_clinical. One generated scan is used as the annotated scans for novel tracts, and 16 scans are used as testing set.  

\begin{figure}[!t]
  \centering
  \includegraphics[width=0.99\columnwidth]{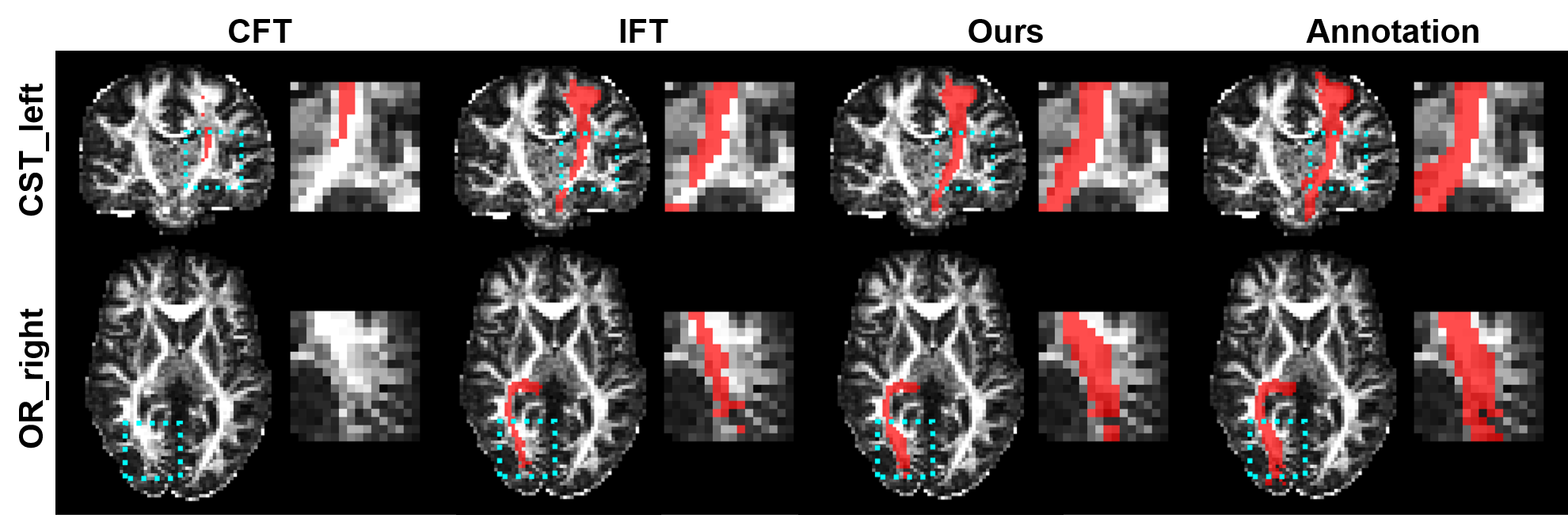}
  \caption{Examples of the segmentation results (red) for novel WM tracts. The results were  obtained in the case of CQ1 and are overlaid on the fractional anisotropy maps. The annotations are also shown for reference. Note the highlighted regions and their zoomed views for comparison.}
%   \caption{Cross-sectional views of the segmentation results (red) overlaid on the fractional anisotropy maps. Note the highlighted regions and their zoomed views for comparison. The annotations are also shown for reference.}
  \label{fig:compare_fig}
\end{figure}

\subsection{Evaluation of Segmentation Results}

We compared the proposed method with two competing methods, which are the classic fine-tuning strategy and the improved fine-tuning strategy~\cite{lu2021knowledge} described in Section~\ref{sec:tl} with the same pretrained model and single annotated training scan used by the proposed method.
Both competing methods were integrated with TractSeg~\cite{wasserthal2018tractseg} like the proposed method.
For convenience, the classic fine-tuning strategy and the improved fine-tuning strategy are referred to as CFT and IFT, respectively.
Note that as shown in~\cite{lu2021knowledge}, in the one-shot setting directly training a model that segments novel WM tracts from scratch without the pretrained model would lead to segmentation failure. Thus, this strategy was not considered.

% First, we qualitatively evaluated the proposed method and compared its segmentation results with those of the competing methods.
We first qualitatively evaluated the proposed method.
Examples of the segmentation results for novel WM tracts are shown in Fig.~\ref{fig:compare_fig} for the proposed and competing methods, where the annotations are also displayed for reference. 
For demonstration, here we show the results obtained in the case of CQ1 for CST\_left and OR\_right. We can see that the segmentation results of our method better resemble the annotations than those of the competing methods.

% Next, we quantitatively compared our method with the competing methods, where the Dice coefficient was computed for the segmentation result of each novel WM tract on each scan for each case.
Next, we quantitatively evaluated our method. The Dice coefficient was computed for the segmentation result of each novel WM tract on each test scan for each case.
For demonstration, we have listed the average Dice coefficient of each novel WM tract in Table~\ref{tab:quantitative_results} for the cases of CQ1 and IH1.
%For demonstration, we have listed the average Dice coefficient of each novel WM tract in Table~\ref{tab:quantitative_results} for the cases of CQ1 and IH1, and the average Dice coefficients for the other cases are given in Table~S1 in the supplementary materials.
For each tract and each case, our method has a higher average Dice coefficient than the competing methods, and the improvement is statistically significant.
We have also summarized the mean of the average Dice coefficients of the novel WM tracts for all the cases in Table~\ref{tab:quantitative_results_mean} (upper half table). In all cases our method outperforms the competing methods with higher mean Dice coefficients.

\begin{table}[!t]
\caption{The average Dice coefficient of each novel WM tract for each method for the cases of CQ1 and IH1. The best results are highlighted in bold. Asterisks indicate that the difference between the proposed method and the competing method is significant using a paired Student's $t$-test. ($^{***}p<0.001$)} 
\label{tab:quantitative_results}
    \centering
    \resizebox{0.99\columnwidth}{!}{
    \begin{tabular}{>{\arraybackslash}p{2cm}
    >{\centering\arraybackslash}p{1cm} >{\centering\arraybackslash}p{1cm} >{\centering\arraybackslash}p{1cm}
    >{\centering\arraybackslash}p{1cm} >{\centering\arraybackslash}p{1cm}
    >{\centering\arraybackslash}p{1cm}
    >{\centering\arraybackslash}p{1cm} >{\centering\arraybackslash}p{1cm}
    >{\centering\arraybackslash}p{1cm} >{\centering\arraybackslash}p{1cm}}
        \hline
        \hline
        &\multicolumn{5}{c}{CQ1}&\multicolumn{5}{c}{IH1}\\
        % \hline
        \cmidrule(lr){2-11}
        WM Tract & \multicolumn{2}{c}{CFT} & \multicolumn{2}{c}{IFT} & Ours & \multicolumn{2}{c}{CFT} & \multicolumn{2}{c}{IFT} & Ours\\% \multirow{2}{*}{Method} & \multicolumn{2}{c}{CFT} & \multicolumn{2}{c}{IFT~\cite{lu2021knowledge}} & Ours & \multicolumn{2}{c}{CFT} & \multicolumn{2}{c}{IFT~\cite{lu2021knowledge}} & Ours\\
        \cmidrule(lr){2-6}
        \cmidrule(lr){7-11}
        &Dice&$p$&Dice&$p$&Dice&Dice&$p$&Dice&$p$&Dice\\
        \hline
        CST\_left&0.123&***&0.463&***&\textbf{0.644}&0.208&***&0.455&***&\textbf{0.569}\\
        CST\_right&0.120&***&0.564&***&\textbf{0.692}&0.129&***&0.416&***&\textbf{0.542}\\
        OR\_left&0.000&***&0.281&***&\textbf{0.492}&0.118&***&0.504&***&\textbf{0.548}\\
        OR\_right&0.000&***&0.401&***&\textbf{0.533}&0.100&***&0.462&***&\textbf{0.518}\\
        \hline
        \hline
    \end{tabular}}
\end{table}

\begin{table}[!t]
\caption{The mean value of the average Dice coefficients of the novel WM tracts for each case and each method. The best results are highlighted in bold. The results achieved with each individual data augmentation strategy in our method are also listed.}
% Asterisks indicate that the difference between the proposed method and the competing method is significant using a paired Student's $t$-test. ($^{*}p<0.05$, $^{**}p<0.01$, $^{***}p<0.001$, n.s. $p>0.05$)}
\label{tab:quantitative_results_mean}
    \centering
    \resizebox{0.93\columnwidth}{!}{
    \begin{tabular}{>{\centering\arraybackslash}p{1.6cm} >{\centering\arraybackslash}p{1.6cm} >{\centering\arraybackslash}p{1.6cm}
    >{\centering\arraybackslash}p{1.6cm} >{\centering\arraybackslash}p{1.6cm}
    >{\centering\arraybackslash}p{1.6cm} >{\centering\arraybackslash}p{1.6cm}}
\hline
\hline
Method & CQ1 & CQ2 & CQ3 & IH1 & IH2 & IH3 \\
\hline

CFT &0.061&0.097&0.092&0.139&0.280&0.192\\
IFT &0.427&0.351&0.396&0.459&0.518&0.531\\
Ours&\textbf{0.590}&\textbf{0.611}&\textbf{0.567}&\textbf{0.544}&\textbf{0.639}&\textbf{0.662}\\
\hline
RC1 &0.574&0.552&0.519&0.529&0.630&0.637\\
RC2 &0.552&0.555&0.544&0.514&0.625&0.651\\
TC1 &0.524&0.605&0.536&0.509&0.592&0.637\\
TC2 &0.582&0.610&0.552&0.524&0.629&0.654\\
\hline
\hline
    \end{tabular}}
\end{table}

Finally, we confirmed the benefit of each proposed data augmentation strategy, as well as the benefit of ensembling their results.
For each case, the mean value of the average Dice coefficients of all novel WM tracts was computed for the segmentation results achieved with RC1, RC2, TC1, or TC2 individually, and the results are also given in Table~\ref{tab:quantitative_results_mean} (lower half table).
%For each case, the mean value of the average Dice coefficients of all novel WM tracts was computed for the segmentation results achieved with RC1, RC2, TC1, or TC2 separately, and the results are also given in Table~\ref{tab:quantitative_results_mean} (lower half table).
Compared with the results of IFT that did not use the proposed data augmentation, the integration of IFT with RC1, RC2, TC1, or TC2 led to improved segmentation accuracy, which indicates the individual benefit of each proposed data augmentation strategy.
% Compared with the results of IFT that did not use the proposed data augmentation, each proposed data augmentation strategy integrated with IFT led to improved segmentation accuracy, which indicates the individual benefit of each data augmentation strategy.
In addition, the Dice coefficients of the proposed method achieved with ensembling are higher than those achieved with a single data augmentation strategy, which confirms the benefit of ensembling.
Note that there is not a data augmentation strategy that is better or worse than the others in all cases, which is possibly because of the randomness in RC1 and RC2 and the dependence of TC1 and TC2 on the spatial coverages of the novel WM tracts that vary in different cases.

\section{Conclusion}

We have proposed an approach to one-shot segmentation of novel WM tracts based on an improved pretraining and fine-tuning framework via extensive data augmentation.
The data augmentation is performed with region masking, and several masking strategies are developed. The segmentation results achieved with these strategies are ensembled for the final segmentation.
% we extend Cutout to our task and utilize the random 3D box or the distribution information of novel WM tracts to modify the input image and tract annotation of the single annotated subject. 
% The segmentation results using different data augmentation strategies are ensembled to obtain the final segmentation results.
The experimental results on two brain dMRI datasets show that the proposed method improves the accuracy of novel WM tract segmentation in the one-shot setting.
% ~{\color{red}It is still possible to further improve the performance on clinical datasets with even more diverse data augmentation, which can be investigated in future work.}

\subsubsection{Acknowledgements}
This work is supported by Beijing Natural Science Foundation (L192058).

\bibliographystyle{splncs04}
\bibliography{refs}

\end{document}